\documentclass[10pt,twocolumn,letterpaper]{article}

\usepackage[pagenumbers]{cvpr} % To force page numbers, e.g. for an arXiv version

\makeatletter
\@namedef{ver@everyshi.sty}{}
\makeatother

\usepackage[accsupp]{axessibility}
\usepackage{graphicx}
\usepackage{amsmath}
\usepackage{amssymb}
\usepackage{booktabs}
\usepackage{csquotes}
\usepackage{enumitem}
\usepackage[acronym]{glossaries}
\usepackage{tabularx}
\usepackage{multicol}
\usepackage{multirow}
\usepackage{makecell}
\usepackage{textcomp}
\usepackage{gensymb}
\usepackage{tikz}
\usepackage[T1]{fontenc}

\usepackage[pagebackref,breaklinks,colorlinks]{hyperref}

\newcommand{\shortsec}[1]{\noindent{\textbf{#1}}~}
\def\shname{GaitGraph2}

\usepackage[capitalize]{cleveref}
\crefname{section}{Sec.}{Secs.}
\Crefname{section}{Section}{Sections}
\Crefname{table}{Table}{Tables}
\crefname{table}{Tab.}{Tabs.}

\def\cvprPaperID{1} % *** Enter the CVPR Paper ID here
\def\confName{BioMetCVPR}
\def\confYear{2022}

\newacronym{cnn}{CNN}{Convolutional Neural Network}
\newacronym{3d-cnn}{3D-CNN}{3D Convolutional Neural Network}
\newacronym{gcn}{GCN}{Graph Convolutional Network}
\newacronym{gei}{GEI}{Gait Energy Image}
\newacronym{cvae}{CVAE}{Conditional Variational Autoencoder}
\newacronym{sota}{SotA}{state-of-the-art}
\newacronym{ptsn}{PTSN}{Pose-Based Temporal-Spatial Network}
\newacronym{lstm}{LSTM}{Long Short-Term Memory}
\newacronym{stgcn}{ST-GCN}{Spatial-Temporal Graph Convolutional Network}
\newacronym{tta}{TTA}{Test Time Augmentation}

\begin{document}

\title{Towards a Deeper Understanding of Skeleton-based Gait Recognition}

\author{
Torben Teepe$^*$ \qquad Johannes Gilg \qquad Fabian Herzog  \qquad Stefan H\"ormann \qquad Gerhard Rigoll\vspace{5pt}\\
Technical University of Munich\\
  {\tt\small $^*$t.teepe@tum.de}\\
}

\maketitle

%%%%%%%%% ABSTRACT
\begin{abstract}
Gait recognition is a promising biometric with unique properties for identifying individuals from a long distance by their walking patterns.
In recent years, most gait recognition methods used the person's silhouette to extract the gait features.
However, silhouette images can lose fine-grained spatial information, suffer from (self) occlusion, and be challenging to obtain in real-world scenarios.
Furthermore, these silhouettes also contain other visual clues that are not actual gait features and can be used for identification, but also to fool the system.
Model-based methods do not suffer from these problems and are able to represent the temporal motion of body joints, which are actual gait features.
The advances in human pose estimation started a new era for model-based gait recognition with skeleton-based gait recognition.
In this work, we propose an approach based on \glspl{gcn} that combines higher-order inputs, and residual networks to an efficient architecture for gait recognition.
Extensive experiments on the two popular gait datasets, \mbox{CASIA-B} and OUMVLP-Pose, 
show a massive improvement (3$\times$) of the \gls{sota} on the largest gait dataset OUMVLP-Pose and strong temporal modeling capabilities. 
Finally, we visualize our method to understand skeleton-based gait recognition better and to show that we model real gait features.
\end{abstract}

%%%%%%%%% BODY TEXT
\section{Introduction}\label{sec:introduction}
Gait is a soft biometric with huge, unique advantages compared to hard biometrics like face, iris, or fingerprint. Human gait can be described as the way a person walks, or more formal, the movement pattern of the limbs during motion.
Gait patterns can be observed at a distance, without a person's compliance, and are hard to disguise. This is a considerable advantage compared to hard biometrics, which requires the user to interact with a sensor. For applications like surveillance and forensic identification, gait offers vast potential; however, it also entails risks for privacy and misuse.

\begin{figure}[t]
    \centering
    \includegraphics[width=.7\columnwidth]{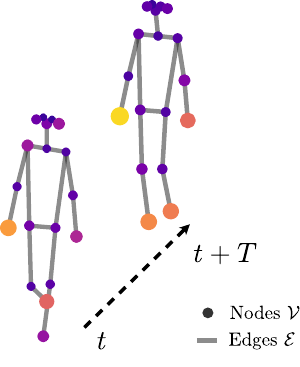}
    \caption{Example of two poses represented as graph at different time steps.}
    \label{fig:gait_math}
\end{figure}

There are also challenges when using gait as a biometric. Gait is sensitive to carried items, worn cloth, and surface type.
A different type of footwear like sneakers compared to boots or heels may considerably change a person's gait.
The biggest challenge in gait recognition is obtaining unique features invariant to these influences.

\begin{figure}[t]
  \centering
  \includegraphics[width=\columnwidth]{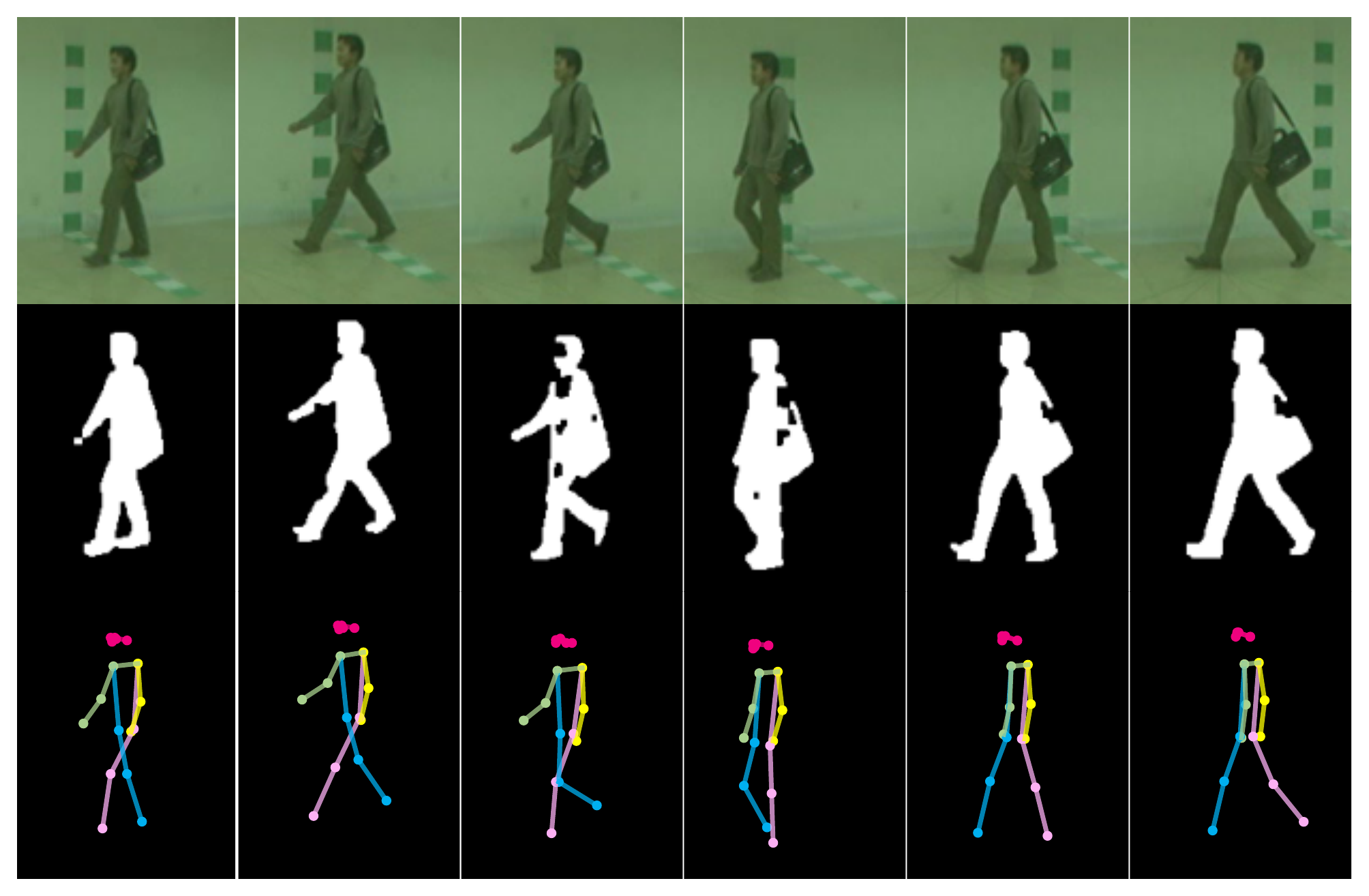}
  \caption{Comparison a sequence of RGB images and the respective gait representation in silhouette image and skeleton, from top to bottom. Images are from the \mbox{CASIA-B}~\cite{yu2006framework} dataset.}
  \label{fig:gait_representation}
\end{figure}

Most approaches \cite{wang2003silhouette, han2005individual, wolf2016multi, feng2016learning, babaee2019gait, song2019gaitnet, chao2019gaitset, fan2020gaitpart} use silhouettes to extract the gait features from a video sequence.
The silhouette is commonly obtained by background subtraction. Depending on the method used for background subtraction, this may cause undesired artifacts on the contour, as shown in the two center frames in \cref{fig:gait_representation}.
Background subtraction may be reliable in a lab setting but becomes a complex problem in a real-world scenario with cluttered and rapidly changing environments.
While most approaches \cite{wang2003silhouette, han2005individual, wolf2016multi, feng2016learning, babaee2019gait, chao2019gaitset, fan2020gaitpart} do not consider how to obtain the silhouettes, one approach \cite{song2019gaitnet} trains a separate \gls{cnn} for background subtraction in a real-world scenario.
A significant drawback of the silhouette representation is the sensitivity to clothing and carried items.
In \cref{fig:gait_representation} the persons bag is clearly visible in the silhouette.
Furthermore, the silhouette reveals many appearance clues, like hairstyle and clothes.
Hence, these approaches are more comparable to re-identification tasks.

Recent deep learning based pose estimation allow to generate keypoints robust against occlusion, cluttered, and changing backgrounds~\cite{sun2019deep}.
These advantages make an approximation to gait features with appearance-based methods obsolete and enable a new generation of model-based gait recognition \cite{sokolova2018pose, liao2017pose, liao2020model}.
Building upon keypoints brings gait recognition back to an early description of gait \cite{johansson1973visual}:
\begin{displayquote}
A few bright spots describing the motions of the main joints [...] evoke a compelling impression of human walking.
\end{displayquote}
Skeleton-based approaches offer a cleaner gait representation, only capturing the spatial posture  and the temporal movement.
Thus we can bring back \textit{actual} gait with a focus on motion recognition instead of visual recognition. 

Current model-based approaches \cite{sokolova2018pose, liao2017pose, liao2020model} still lack performance compared to appearance-based methods.
We introduce \glspl{gcn} to process the skeletons described by the keypoints and bridge the gap to appearance-based methods even further. \\
Our contributions can be summarized as follows:
\begin{enumerate}[nosep]
\item We propose a multi-branch graph-based interpretation of gait together with a \gls{gcn} architecture that can efficiently learn features on this graph. 
\item We provide a deeper understanding of gait with extensive ablation and visualization of our features.
\item Our empirical experiments show \gls{sota} results by a huge margin on the largest model-based gait dataset OUMVLP-Pose.
\end{enumerate}
The \href{https://github.com/tteepe/GaitGraph2}{code and models} are publicly available\footnote{\href{https://github.com/tteepe/GaitGraph2}{\tt https://github.com/tteepe/GaitGraph2}}.

% Code and models are available at: \footnote{Code will be made publicly available}.

\section{Related Work}
Although skeleton-based gait recognition is a relatively new research area, silhouette-based approaches have a long history. This section will give an overview of these two areas of gait recognition and other skeleton-based human understanding that inspired this work.

\subsection{Gait Recognition}
In recent years, appearance-based approaches using a silhouette representation as input dominated gait recognition. Model-based approaches only played a minor role, but lately, new model-based approaches have emerged using pose estimation and the human skeleton as the gait feature representation. Silhouette-based methods still set the \gls{sota} for gait recognition, but recently skeleton-based approaches have started to challenge this lead.

\subsubsection{Silhouette-based}
Silhouette-based methods relied on a binary human image extracted from the original image \cite{wang2003silhouette}. Background subtraction can obtain these silhouettes for static scenes, but for dynamic and changing settings, this task becomes more complicated \cite{song2019gaitnet}.

Silhouette approaches are distinctive for their temporal modeling and group into single-image, sequence-based, and set-based approaches. Early approaches summarized a gait cycle into a single image, i.e., \gls{gei} \cite{han2005individual, babaee2019gait}. These representations lose most of the temporal information but allow for easier processing. Sequence-based approaches focus on each input separately. For modeling the temporal information 3D-\gls{cnn}~\cite{wolf2016multi} or \glspl{lstm}~\cite{feng2016learning} are used. These approaches can comprehend more spatial information and more temporal information at higher computational costs. The set-based approach \cite{chao2019gaitset, fan2020gaitpart} with shuffled inputs models no temporal information, thus has less computational complexity.

\subsubsection{Skeleton-based}
Model-based approaches were not very popular in recent years due to the high computing complexity. 
Robuster and more lightweight pose estimators \cite{cao2019open, fang2017rmpe} enabled a comeback for model-based approaches. The keypoints of the human body allow modeling a skeleton representation of a person. The first approach to propose the use of pose estimation and utilize pose keypoints is \gls{ptsn}~\cite{liao2017pose}. \textit{\gls{ptsn}} proposed a two-path network architecture with an \gls{cnn} for spatial modeling, and an \gls{lstm} for temporal modeling. The authors show superior results on CASIA-B~\cite{yu2006framework} compared to silhouette-based approaches in special walking conditions in the same-view setting.

Later, the same group proposed \textit{PoseGait}~\cite{liao2020model}, which uses 3D pose estimation with handcrafted features. The approach uses the 3D keypoints in euclidean space to calculate the joint angle, bone length, and joint motion. Using these handcrafted features, a \gls{cnn} then learns high-level spatio-temporal features. This approach is evaluated in a cross-view setting on CASIA-B~\cite{yu2006framework} and shows competitive results to silhouette-based approaches.

We build on the work GaitGraph \cite{teepe2021gaitgraph}, which introduced the first approach with \glspl{gcn} for skeleton-based gait recognition. With an adapted \gls{gcn} architecture for action recognition and integrated spatial and temporal modeling, GaitGraph gave a considerable performance increase for model-based gait recognition.

A recent approach \cite{wang2022twobranch} combines the advantages of silhouette-based and skeleton-based approaches in a two-branch \gls{gcn} and \gls{cnn} network architecture and achieve even higher recognition rates. Showing that skeleton-based approaches preserve information not captured in silhouette-based methods.

\subsection{Skeleton-based Action Recognition}
While skeleton-based gait recognition was only recently becoming popular, other areas of human understanding have already employed skeleton features for a few years.
First and foremost, the area of skeleton-based action recognition pioneered most of the current graph-based spatio-temporal \gls{gcn} architectures, including the ones used in this paper \cite{yan2018spatial, song2020stronger}.

Yan \etal \cite{yan2018spatial} introduced \glspl{gcn} to action recognition with the \gls{stgcn}
architecture, in which skeleton data is represented as a graph with natural skeleton connections. \gls{stgcn} interleaves spatial graph convolutions along with temporal convolutions for spatial-temporal modelling. Many more \gls{gcn} architectures and improvements on the original \gls{stgcn} design have since been proposed. A notable example is the Two-Stream Adaptive Graph Convolutional Network (2s-AGCN) \cite{shi2019two} that introduced a adaptive adjacency matrix and pre-computed bone information as second-order input.
SGN \cite{zhang2020semantics} provides a deeper analysis of these second-order features and proposes a data pre-processing step that adds pre-computed velocity and bone information to the raw keypoint input. 
Another architecture used in this paper is ResGCN \cite{song2020stronger}, which added multiple residual connections in the ST-GCN blocks and a bottleneck for feature dimension reduction. ResGCN also adopts the higher-order input with a multi-branch input structure.

\section{Skeleton-based Gait Recognition}

\subsection{Notation}
We describe the human skeleton as a graph $\mathcal{G} = (\mathcal{V}, \mathcal{E})$, where $\mathcal{V} = \{v_1, \hdots, v_N\}$ is the set of $N$ nodes representing joints, and $\mathcal{E}$ is the set of edges representing bones captured by an adjacency matrix $\mathbf{A} \in \mathbb{R}^{N \times N}$ with $\mathbf{A}_{i,j} = 1$ if an edge connects from $v_i$ to $v_j$ and $\mathbf{A}_{i,j} = 0$ otherwise.
$\mathbf{A}$ is symmetric since $\mathcal{G}$ is undirected.
Every node consists of three channels $v_n = (x_n, y_n, c_n)$, with the estimated $x,y$ coordinate and the keypoint confidence $c$.

For gait recognition, we use a sequence of these graphs.
Thus we add a temporal dimension $T$.
The sequence is then defined as the tensor $\mathbf{X} = \{v_{t,n} \in \mathbb{R}^{3} \mid t,n \in \mathbb{N}_0,t < T, n < N\}$ and $\mathbf{X} \in \mathbb{R}^{T \times N \times C}$.

Overall the gait sequence can be described structurally by the adjacency matrix $\mathbf{A}$ and the feature tensor $\mathbf{X}$.

\subsection{Graph Convolutions}
An essential building block of our network architecture are graph convolutions. On skeleton inputs, defined by features $\mathbf{X}$ and graph structure $\mathbf{A}$, the layer-wise update rule of graph convolutions can be applied to features at time $t$ as:
\begin{equation} \label{eq:gcn}
    \mathbf{X}_t^{(l+1)} = \sigma\left(
        \tilde{\mathbf{D}}^{-\frac{1}{2}}
        \tilde{\mathbf{A}}
        \tilde{\mathbf{D}}^{-\frac{1}{2}}
        \mathbf{X}_t^{(l)}
        \Theta^{(l)}
    \right),
\end{equation}
where $\tilde{\mathbf{A}} = \mathbf{A + I}$ is the skeleton graph with added self-loops to keep identity features. $\tilde{\mathbf{D}}$ is the diagonal degree matrix of $\tilde{\mathbf{A}}$, and $\sigma(\cdot)$ is an activation function.
The term
$
\tilde{\mathbf{D}}^{-\frac{1}{2}}
\tilde{\mathbf{A}}
\tilde{\mathbf{D}}^{-\frac{1}{2}}
\mathbf{X}_t^{(l)}
$
can be intuitively interpreted as a spatial feature aggregation from the messages passed by the direct neighbors.
The adjacency matrix $\mathbf{A}$ is obtained using the spatial partition presented in \cite{yan2018spatial}.
% maybe explain how it works

\subsection{Pose Estimation}
The skeletons are extracted from the RGB images of the dataset.v Keypoint estimation aims to detect the locations of $N$ keypoints (e.g., shoulder, hip, or knee) from an image $\mathbf{I} \in \mathbb{R}^{W \times H \times 3}$.
A common method is the top-down-approach \cite{sun2019deep}, which predicts $N$ heatmaps $\left\{\mathbf{H}_1, \mathbf{H}_2, \ldots, \mathbf{H}_N \right\}$ of size $W' \times H'$, where the heatmap $\mathbf{H}_n$ indicates the location of the $n$-th keypoint.
The location of the maximum of these heatmaps $\mathbf{H}_n$ yields the location of the keypoint $v_n$ that define the edges $\mathcal{E}$.

Official keypoints \cite{an2020performance} for the OUMVLP dataset keypoints are provided.
The authors rely real-time pose estimators to allow real-time applications of the approach.
In our work, we decided to extract the 2D keypoints on CASIA-B using HRNet \cite{sun2019deep} pre-trained on the COCO dataset \cite{lin2014microsoft}.
It is an offline approach but yields higher keypoint accuracy than real-time approaches.
\cref{tab:datasets} shows a comparison of the two datasets and the used keypoints.
The COCO pose annotations consist of 17 keypoints. We define the bones or edges $\mathcal{E}$ as shown in \cref{fig:gait_math}. 

\subsection{Network Architecture}
For our task, we adapted the ResGCN \cite{song2020stronger} architecture designed initially for action recognition.
Blocks of this architecture are based on the ST-GCN block. and contain a sequential execution of a spatial graph convolutions and a temporal 2D convolutions.
The ResGCN approach introduces a bottleneck structure, based on ResNet by He \etal \cite{he2016deep}.
The bottleneck adds two 1 $\times$ 1 convolutional layers before and after a convolution layer to reduce the number of feature channels with a reduction rate $r$.
ResGCN adds this bottleneck for every spatial and temporal block, thus reducing the number of parameters.

Another modification to the original ST-GCN architecture is the addition of residual connections. 
Song \etal \cite{song2020stronger} propose residual connections that connect the features before and after every spatial and temporal block.

\begin{figure}[t]
    \centering
    \includegraphics[width=0.45\columnwidth]{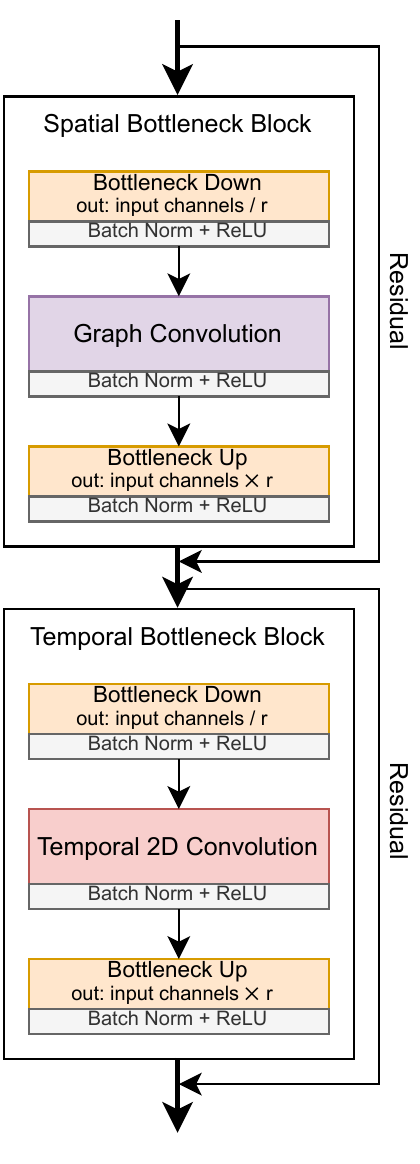}
    \caption{Structure of ResGCN bottleneck block including the residual connections.}
    \label{fig:resgcn-block}
\end{figure}

\begin{figure*}[t]
    \centering
    \includegraphics[width=0.99\textwidth]{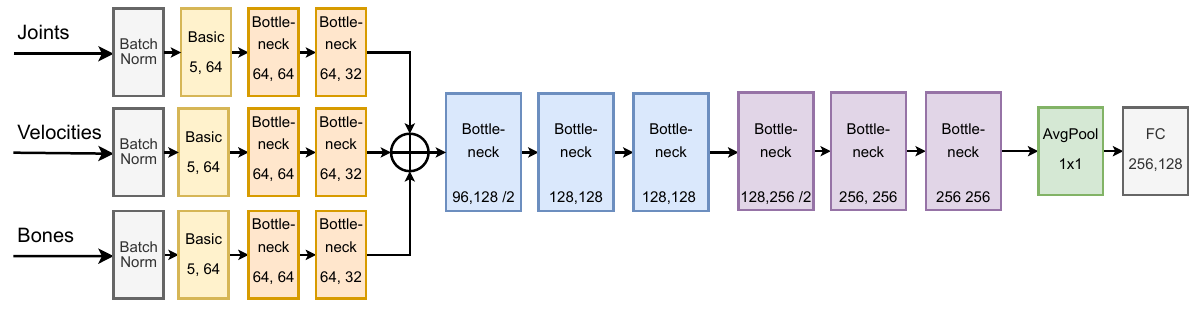}
    \caption{Overview of the multi-branch ResGCN architecture.}
    \label{fig:resgcn}
  \end{figure*}

A ResGCN bottleneck block is shown in \cref{fig:resgcn-block} and an overview of the overall structure in \cref{fig:resgcn}. 

\subsection{Multi-Branch Input}
As proposed by works in skeleton-based gait recognition \cite{zhang2020semantics, song2020stronger}, we pre-compute features of our skeleton information.
In this work, we use three features: 1) joint positions, 2) motion velocities, and 3) bone features.

The first branch contains the joint positions. We also add the relative position of each joint to the center of the pose $c$ (\ie nose or neck):
$$\mathcal{R} = \{v_{t,n} - v_{t,c} \mid t,n \in \mathbb{N}_0, t < T,  n < N\}.$$

The second branch uses the motion velocities $\mathcal{F}$ as input. We calculate the difference to the same joint in the two next frames for each joint:
$$\mathcal{F} = \{v_{t + i,n} - v_{t,n} \mid t,n \in \mathbb{N}_0, i \in \{1,2\}, t < T - 2, n < N\}.$$

Finally, the input of the last branch is the bone length $\mathcal{L}$ and bone angles $\mathcal{A}$.
For the bone length, we subtract the coordinate of every joint $n$ with every connected joint $n_{adj}$:
$$\mathcal{L} = \{v_{t,n} - v_{t,n_adj} \mid t,n \in \mathbb{N}_0, t < T,  n < N\}.$$
Finally, the angle of each bone is:
\begin{small}
$$\small\mathcal{A} = \left\{\arccos{\left(\frac{v_{t,n} - v_{t,n_adj}}{\sqrt{\sum{v_{t,n}^2}}}\right)} \mid t,n \in \mathbb{N}_0, t < T,  n < N\right\}.$$
\end{small}

\section{Experiments}

\subsection{Datasets}
In recent years, the focus for gait recognition was on silhouette-based approaches. Hence, most datasets provided only silhouettes. One of the widely used datasets, CASIA-B \cite{yu2006framework}, provides RGB images on which we can run the pose estimation. Recently, an extension with the pose data OUMVLP-Pose \cite{an2020performance} was published for the largest public gait database \mbox{OU-MVLP} \cite{takemura2018multi}. \cref{tab:datasets} shows a comparison of the two datasets. These two popular gait datasets provide a good comparison to other methods.

\textbf{CASIA-B} \cite{yu2006framework} is popular multi-view gait dataset with 124 subjects. The dataset contains 3 walking conditions recorded in 11 views (0\degree, 18\degree, \dots, 180\degree). The walking conditions are normal (NM) (6 sequences per subject), walking with a bag (BG) (2 sequences per subject), and wearing a jacket or a coat (CL) (2 sequences per subject). In total each subject contains $11\times(6+2+2)=110$ sequences.

CASIA-B has no official partition of training and test set, but several experiment protocols exist \cite{zhang2019gait}. In this work, we use the popular protocol proposed in \cite{wu2017comprehensive} for a fair comparison. 
Furthermore, we use the commonly called large-sample training (LT) partition. The train set contains the first 74 subjects for this partition, and the remaining 50 subjects build the test set. In the test set, the gallery comprises four sequences of the NM condition (NM \#1-4), and the remaining six sequences are divided into three probe subsets, i.e., NM subsets containing NM \#5-6, BG subsets containing BG \#1-2 and CL subsets containing CL \#1-2.\\
We extract the poses from CASIA-B with the pre-trained HRNet~\cite{sun2019deep} pose estimator.

\textbf{OUMVLP-Pose} \cite{an2020performance} is based on the multi-view large-scale gait dataset, OUMVLP \cite{takemura2018multi}. OUMVLP is currently the largest gait dataset and contains 10,307 subjects captured by seven cameras in a round-trip walking course, resulting in effectively 14 views in a 15\degree\, interval (0\degree, 15\degree, \dots, 90\degree; 180\degree, 195\degree, \dots, 270\degree). Every sequence contains from 18 to 35 frames, and on average, 25 frames. The dataset is split into 5,153 subjects for training and 5,154 subjects for testing. For testing, sequences with index \#01 assemble the gallery, while the other sequences are used for the probe set.

OUMVLP-Pose keypoints are extracted from the (unreleased) RGB frames. Two datasets were created using different pre-trained pose estimators, OpenPose \cite{cao2019open}, and AlphaPose \cite{fang2017rmpe}, but containing the same frames and subjects. 

\begin{table}
   \fontsize{9}{11}\selectfont
    \centering
    
    \begin{tabular}{r | c | c | c }
    \toprule
        & CASIA-B & \multicolumn{2}{c}{OUMVLP-Pose} \\
    \midrule
        \#IDs & 124 & \multicolumn{2}{c}{10,307}\\
        \#Sequences per ID & 2 $\times$ 3  & \multicolumn{2}{c}{2}\\
        Keypoint Estimatior & HRNet & OpenPose & AlphaPose\\
        Keypoint Accuracies & 75.8 mAP & 64.2 mAP & 71.0 mAP\\
        \#Keypoints& 17 & \multicolumn{2}{c}{18}\\
    \bottomrule
   \end{tabular}
   \caption{Comparison of the two skeleton gait datasets. Keypoint accuracy of the pose estimator is reported on the COCO test-dev dataset.}
   \label{tab:datasets}
   \end{table}

\subsection{Implementation Details}
For the training setup, we use an Adam optimizer with a \textit{1-cycle} learning rate schedule \cite{smith2019super} with a maximum learning rate of 0.005.
% more about one-cycle config
The embedding size is 128, the loss function's temperature is 0.01, and the batch size is 768.
After 80\% of the maximum epochs, we use Stochastic Weight Averaging (SWA)~\cite{izmailov2018averaging}.
All the experiments are conducted on a single NVIDIA 3090 GPU with PyTorch.
Due to the different properties of the size of the datasets, we employ different model sizes and training strategies.

For \textbf{CASIA-B} we use the \textit{ResGCN-N21-R8} architecture with 350 K parameters and a sequence length of $T=60$ and train for 200 epochs.

For \textbf{OUMVLP-Pose} we us a sequence length of $T=30$ and train for 750 epochs. The network setup is \textit{ResGCN-N51-R4} with 765 K parameters.

% For the experiments with the ST-GCN model, the dropout layer in the original ST-GCN model \cite{yan2018spatial} is removed.

\shortsec{Loss}
As the loss function, we use supervised contrastive (\textit{SupCon}) loss as proposed by \cite{khosla2020supervised}.
% The loss is computed with the folling equation: 
% \begin{equation}
%     \mathcal{L}_{out}^{sup}
%     =\sum_{i\in I}\frac{-1}{|P(i)|}\sum_{p\in P(i)}\log{\frac{\text{exp}\left(\boldsymbol{z}_i\cdot\boldsymbol{z}_p/\tau\right)}{\sum\limits_{a\in A(i)}\text{exp}\left(\boldsymbol{z}_i\cdot\boldsymbol{z}_a/\tau\right)}}
%     \label{eqn:supervised_loss}
%   \end{equation}
Compared to traditional contrastive losses such as triplet loss or N-pairs loss, this current batch contrastive loss considers all positive and negative samples in the batch. The small size of the skeleton data allows us to run big batch sizes; thus, each batch should contain a positive pair. If an element has only negative pairs or no pairs, it will be ignored.

\shortsec{Augmentation}
Our network design relies on a constant input sequence length. Thus, we first add mirror padded frames for sequences shorter than the desired sequence length in the temporal dimension. Afterward, we randomly pick a subsequence of the desired length.
Additionally, we flip the images from left to right and flip every left joint with the right and vice versa. We add various uniform noises to the estimated keypoints and their confidence to account for pose estimator inaccuracies. 

\shortsec{Evaluation \& \gls{tta}}
At test time, the distance between gallery and probe is defined as the cosine similarity of the corresponding feature vectors.
The same sequence padding augmentation from train time is applied, but we pick a sub-sequence of the required length from the sequence center.
We also use two additional samples: a left/right flipped sample and a time inverted sample. The resulting three embeddings are concatenated for the later distance calculation.

\begin{table*}[tbp]
\centering
% \fontsize{9}{11}\selectfont
\begin{tabular}{l|r|c c c c c c c c c c c|c}
     \toprule
     \multicolumn{2}{l|}{Gallery NM\#1-4}&\multicolumn{11}{c|}{0\degree-180\degree}&\multirow{2}{*}{mean}\\[0.3mm]

     \multicolumn{2}{l|}{Probe}& 0\degree&18\degree&36\degree&54\degree&72\degree&90\degree&108\degree&126\degree&144\degree&162\degree&180\degree\\
     
     \midrule
     
     \multirow{3}{*}{NM\#5-6}
    %  &GaitNet \cite{song2019gaitnet}  & 91.2 & 92.0 &90.5 &95.6 &86.9 &92.6 &93.5 &96.0 &90.9 &88.8 &89.0 &91.6\\
    %  &GaitSet \cite{chao2019gaitset}  & 90.8 & 97.9 &\textbf{99.4} &96.9 &93.6 &91.7 &95.0 &97.8 &98.9 &96.8 &85.8 &95.0\\     
    %  &GaitPart \cite{fan2020gaitpart} & \textbf{94.1} & \textbf{98.6} & 99.3 &\textbf{98.5} &94.0 &92.3 &95.9 & \textbf{98.4} & \textbf{99.2} &\textbf{97.8} &90.4 &\textbf{96.2}\\
     % &PTSN &\\
     &PoseGait \cite{liao2020model}       & 55.3 & 69.6 & 73.9 & 75.0 & 68.0 & 68.2 & 71.1 & 72.9 & 76.1 & 70.4 & 55.4 & 68.7\\
     &GaitGraph \cite{teepe2021gaitgraph} & 85.3 &  88.5 &  91.0 &  92.5 &  87.2 &  86.5 &  88.4 &  89.2 &  87.9 &  85.9 &  81.9 &  87.7 \\
     &\textbf{\shname}                    & 78.5 & 82.9 & 85.8 & 85.6 & 83.1 & 81.5 &  84.3 &  83.2 &  84.2 &  81.6 &  71.8 &  82.0   \\
     
     \midrule
     
     \multirow{3}{*}{BG\#1-2}
    %  &GaitNet \cite{song2019gaitnet} & 83.0 &87.8 &88.3 &93.3 &82.6 &74.8 &89.5 &91.0 &86.1 &81.2 &85.6 &85.7\\
    %  &GaitSet \cite{chao2019gaitset} & 83.8 &91.2 &91.8 &88.8 &\textbf{83.3} &81.0 &84.1 &90.0 &92.2 &\textbf{94.4} &79.0 &87.2\\     
    %  &GaitPart \cite{fan2020gaitpart} & \textbf{89.1} &\textbf{94.8} &\textbf{96.7} &\textbf{95.1} &\textbf{88.3} &\textbf{84.9} &\textbf{89.0} &\textbf{93.5} &\textbf{96.1} &93.8 &\textbf{85.8} &\textbf{91.5}\\
     % &PTSN & \\
     &PoseGait \cite{liao2020model}       & 35.3 & 47.2 & 52.4 & 46.9 & 45.5 & 43.9 & 46.1 & 48.1 & 49.4 & 43.6 & 31.1 & 44.5\\
     &GaitGraph \cite{teepe2021gaitgraph} & 75.8 & 76.7 & 75.9 & 76.1 & 71.4 &  73.9 &  78.0 &  74.7 &  75.4 &  75.4 &  69.2 &  74.8 \\
     &\textbf{\shname}                    & 69.9 & 75.9 & 78.1 & 79.3 & 71.4 & 71.7 &  74.3 &  76.2 &  73.2 &  73.4 &  61.7 &   73.2    \\       
     \midrule
     
     \multirow{3}{*}{CL\#1-2}
    %  &GaitNet \cite{song2019gaitnet}  & 42.1 &58.2 &65.1 &70.7 &68.0 &70.6 &65.3 &69.4 &51.5 &50.1 &36.6 &58.9\\
    %  &GaitSet \cite{chao2019gaitset}  & 61.4 &75.4 &80.7 &77.3 &72.1 &70.1 &71.5 &73.5 &73.5 &68.4 &50.0 &70.4\\
    %  &GaitPart \cite{fan2020gaitpart} & \textbf{70.7} &\textbf{85.5} &\textbf{86.9} &\textbf{83.3} &\textbf{77.1} &\textbf{72.5} &\textbf{76.9} &\textbf{82.2} &\textbf{83.8} &\textbf{80.2} &\textbf{66.5} &\textbf{78.7}\\
     % &PTSN &\\
     &PoseGait \cite{liao2020model}       & 24.3 & 29.7 & 41.3 & 38.8 & 38.2 & 38.5 & 41.6 & 44.9 & 42.2 & 33.4 & 22.5 & 36.0 \\
     &GaitGraph \cite{teepe2021gaitgraph} & 69.6 &  66.1 &  68.8 &  67.2 &  64.5 &  62.0 &  69.5 &  65.6 &  65.7 &  66.1 &  64.3 &  66.3 \\
     &\textbf{\shname}                    &57.1 & 61.1 & 68.9 & 66   & 67.8 & 65.4 &  68.1 &  67.2 &  63.7 &  63.6 &  50.4 &   63.6\\
     \bottomrule
\end{tabular}
\caption{Averaged Rank-1 accuracies in percent on CASIA-B per probe angle excluding identical-view cases compared with other model-based methods.}
\label{tab:casia-b-model}
\end{table*}

\begin{table}[!t]
\renewcommand{\arraystretch}{1.0}
\centering
\setlength{\tabcolsep}{3pt}
% \fontsize{9}{11}\selectfont
\begin{tabular}{r|cc|cc}
\toprule
\multirow{3}{0.75cm}{Probe} & \multicolumn{4}{c}{Gallery (0\degree - 270\degree)}\\
      & \multicolumn{2}{c|}{OpenPose} & \multicolumn{2}{c}{AlphaPose}\\
    %   & \multicolumn{2}{c|}{RGB} & \multicolumn{2}{c}{AlphaPose}\\
      & CNN-Pose & \textbf{\shname} & CNN-Pose & \textbf{\shname}\\
\midrule
0\degree    &~8.2 & 32.9 &14.3 & 54.3\\
15\degree   &13.9 & 47.7 &22.3 & 68.4\\
30\degree   &18.1 & 53.9 &27.2 & 76.1\\
45\degree   &22.4 & 56.8 &30.0 & 76.8\\
60\degree   &21.3 & 53.9 &28.4 & 71.5\\
75\degree   &18.2 & 54.7 &23.4 & 75.0\\
90\degree   &10.9 & 45.4 &17.2 & 70.1\\
180\degree  &~7.3 & 29.0 &~7.9 & 52.2\\
195\degree  &13.5 & 35.7 &13.6 & 60.6\\
210\degree  &12.0 & 34.3 &15.6 & 57.8\\
225\degree  &20.5 & 44.3 &25.0 & 73.2\\
240\degree  &17.3 & 46.2 &24.1 & 67.8\\
255\degree  &13.7 & 46.4 &20.2 & 70.8\\
270\degree  &~9.4 & 38.4 &16.5 & 65.3\\
\midrule
mean        &14.8 & \textbf{44.3} &20.4 & \textbf{67.1} \\
\bottomrule
\end{tabular}
\caption{Averaged rank-1 accuracies on OUMVLP-Pose.}
\label{tab:oumvlp}
\end{table}

\begin{table*}
% \fontsize{9}{11}\selectfont
% \setlength{\tabcolsep}{4pt}
 \centering
 
 \begin{tabular}{r | r | c c c | c c c c c c}
 \toprule
  & & \multicolumn{3}{c|}{CASIA-B} & \multicolumn{5}{c}{OU-MVLP}\\
   Type &Method &  NM & BG & CL & 0\degree & 30\degree & 60\degree & 90\degree & mean \\
   \midrule
   \multirow{4}{*}{\makecell[r]{appearance-based}}
   &GEINet~\cite{shiraga2016geinet} & - & - & - & 11.4 & 41.5 & 39.5 & 38.9 & 35.8\\
   &GaitNet~\cite{song2019gaitnet} & 91.6 & 85.7 & 58.9 & - & - & - & - & -\\
   &GaitSet~\cite{chao2019gaitset} & 95.0 & 87.2 & 70.4 &79.5 & 89.9 & 88.1 & 87.8 & 87.1\\
   &GaitPart~\cite{fan2020gaitpart}& \textbf{96.2} & \textbf{91.5} & \textbf{78.7} & \textbf{82.6} & \textbf{90.8} & \textbf{89.7} & \textbf{89.5} & \textbf{88.7}\\
   \midrule
   \multirow{3}{*}{\makecell[r]{model-based}}&PoseGait \cite{liao2020model} & 68.7 & 44.5 & 36.0& - & - & - & - & - \\
   &CNN-Pose\cite{an2020performance} & - & - & - & 12.3 & 29.3 & 30.5 & 18.1 & 20.4 \\
   &GaitGraph \cite{teepe2021gaitgraph} & \textbf{87.7} & \textbf{74.8} & \textbf{66.3} & - & - & - & - & - \\
   &\textbf{\shname}            & 82.0 & 73.2 & 63.6 & \textbf{54.3} & \textbf{76.1} & \textbf{71.5} & \textbf{70.1} & \textbf{67.1} \\
   \midrule
   combined &Two-Branch~\cite{wang2022twobranch}& \textbf{97.7} & \textbf{93.8} & \textbf{92.7} & - & - & - & - & - \\
 \bottomrule
\end{tabular}
\caption{Averaged Rank-1 accuracies in percent comparison with both appearance-based and model-based methods. Results for CASIA-B are in the cross-view setup. Results for OU-MVLP in the model-based categories use OUMVLP-Pose AlphaPose keypoints.}
\label{tab:apperance-vs-model}
\end{table*}

\begin{table}
    \setlength{\tabcolsep}{12.5pt}
    % \fontsize{9}{11}\selectfont
     \centering
     
     \begin{tabular}{l | c | c }
     \toprule
         Model & Params & Acc\\
     \midrule
        ST-GCN (Baseline) & 3.3 M & 60.7\\ % m1 v26
    \midrule
        ResGCN-N51        & 645 K & 63.7\\ % m1 v27
        ~$+$ Multi Input  & 765 K & 66.5\\  %m1 v30
        ~$+$ Augmentation & \textquotedbl & 66.6\\
        ~$+$ TTA          & \textquotedbl & 67.1\\
     \bottomrule
    \end{tabular}
    \caption{Ablation study of the model components. All experiments are conducted on OUMVLP-Pose and AlphaPose keypoints.}
    \label{tab:model_ablation}
    \end{table}

\subsection{Comparison with State-of-the-Art}

First, we compare the results of model-based approaches on \textbf{CASIA-B} in \cref{tab:casia-b-model}.
Currently, we can only compare two approaches.
The other skeleton-based approach PTSN \cite{liao2017pose} did not publish the results with the same evaluation protocol (excluding the same view).
We can still compare to a follow-up paper by the same group PoseGait \cite{liao2020model}. \\ All approaches are more robust in the askew angles than in the strict side view, front view, or back view angles (0\degree, 90\degree, 180\degree). 
Another commonality is that all approaches suffer in performance with different walking conditions.
While the drop is more substantial for PoseGait \cite{liao2020model}, the GCN-based approaches can handle these conditions better.

While our approach builds upon GaitGraph, it can only match the results closely.
Due to the different properties of the two datasets, our approach was more optimized towards the larger, more significant dataset OUMVLP.
The main difference in our approach is the multi-branch input; this makes the training much faster and much more stable compared to GaitGraph.

Secondly, the results on \textbf{OUMVLP-Pose} are in \cref{tab:oumvlp}.
For this recently released dataset, we can only compare to the baseline set by the authors CNN-Pose \cite{an2020performance}.
The results show that our GCN-based methods outperform the CNN-based baseline considerably. We can improve the performance by $\sim$3$\times$ on both keypoint types.

\subsection{Comparison with Appearance-based Methods}
Appearance-based methods still archive the best results in gait recognition. Nevertheless, the new skeleton-based approaches help model-based approaches to close this gap. \cref{tab:apperance-vs-model} shows this comparison.

One hope for skeleton-based gait recognition is to be more invariant to the changes in the walking condition presented in CASIA-B.
For now, the results show that this is not the case. Appearance-based methods can handle these conditions better, and their performance drop is less substantial.
A reason could be that the pose estimators are also not as accurate on people wearing coats and bags.

Our approach is the first one to evaluate the two most popular gait datasets and give the first complete comparison to appearance-based methods.
Skeleton-based approaches made a big step towards the SOTA appearance-based methods; however, there is still a significant gap.

It is also notably that the two-branch approach \cite{wang2022twobranch} that combines both paradigms can improve the results in all walking conditions. It shows that both features are complementary to archive overall SOTA performance. 
Especially people wearing a coat are much better recognized with both input modalities.

\subsection{Ablation Studies}
First, we look at the components of our approach in \cref{tab:model_ablation}.
As a baseline, we use the original ST-GCN architecture.
Following, we analyze how much the different components of our approach contribute to the overall performance.

\shortsec{Temporal Modeling}
The network's ability to model temporal information is investigated by training and testing sorted or shuffled sequences. \cref{tab:temp-ablation} shows three configurations. 
For this ablation study, we can not use the multi-branch inputs since these contain pre-computed temporal information and would harm the validity of the ablation. 

Our approach shows a good ability to model temporal features.
The performance drops substantially when trained with sorted sequences and tested with shuffled sequences (row c).
These results further support our claim of bringing back actual temporal features to gait recognition.
Tab~\ref{tab:temp-ablation} also illustrates the spatial modeling abilities in row a. Despite the missing temporal and appearance information, the network can still learn appearance-invariant features of the person's underlying physic.

\shortsec{Pose Estimator}
The OUMVLP-Pose dataset has keypoints extracted by different pose estimators.
The two follow a different approach for keypoint estimation, with AlphaPose being a top-down approach and OpenPose being a bottom-up approach. 
The performance measured in the detector's mean average precision (mAP) also varies. AlphaPose archives a 71.0 mAP and OpenPose a 64.2 mAP on the COCO test-dev dataset.

These two keypoints allow us to analyze how the gait recognition algorithms scale on differently performing pose estimators.
\cref{tab:oumvlp} shows the results on OUMVLP-Pose with the two keypoint estimators.
We archive almost the same performance gain on both keypoint types compared to CNN-Pose. This scaling performance indicates that skeleton-based methods can scale with the pose estimators' performance, potentially allowing better performance in the future with emerging improved pose estimators.

\begin{table}
% \fontsize{9}{11}\selectfont
\setlength{\tabcolsep}{5.5pt}
 \centering
 
 \begin{tabular}{c | c c | c c c | c }
 \toprule
     & \multicolumn{2}{c|}{} & \multicolumn{3}{c|}{CASIA-B} & OU-MVLP \\
     & Train & Test & NM & BG & CL & AlphaPose\\
    \midrule
    a & Shuffle & Sort & 34.7 & 27.0 & 19.0 & 34.1\\
    b & Sort & Sort    & 72.8 & 60.1 & 44.6 & 63.1\\
    c & Sort & Shuffle & 34.3 & 27.9 & 18.5 & 14.5\\
 \bottomrule
\end{tabular}
\caption{Spatio-temporal Study. Control Condition: shuffle/sort the input sequence at train/test phase. Results are rank-1 accuracies averaged in percent. CASIA-B results are in the cross-view setup. All experiments are without the multi-input pre-computation and TTA.}
\label{tab:temp-ablation}
\end{table}
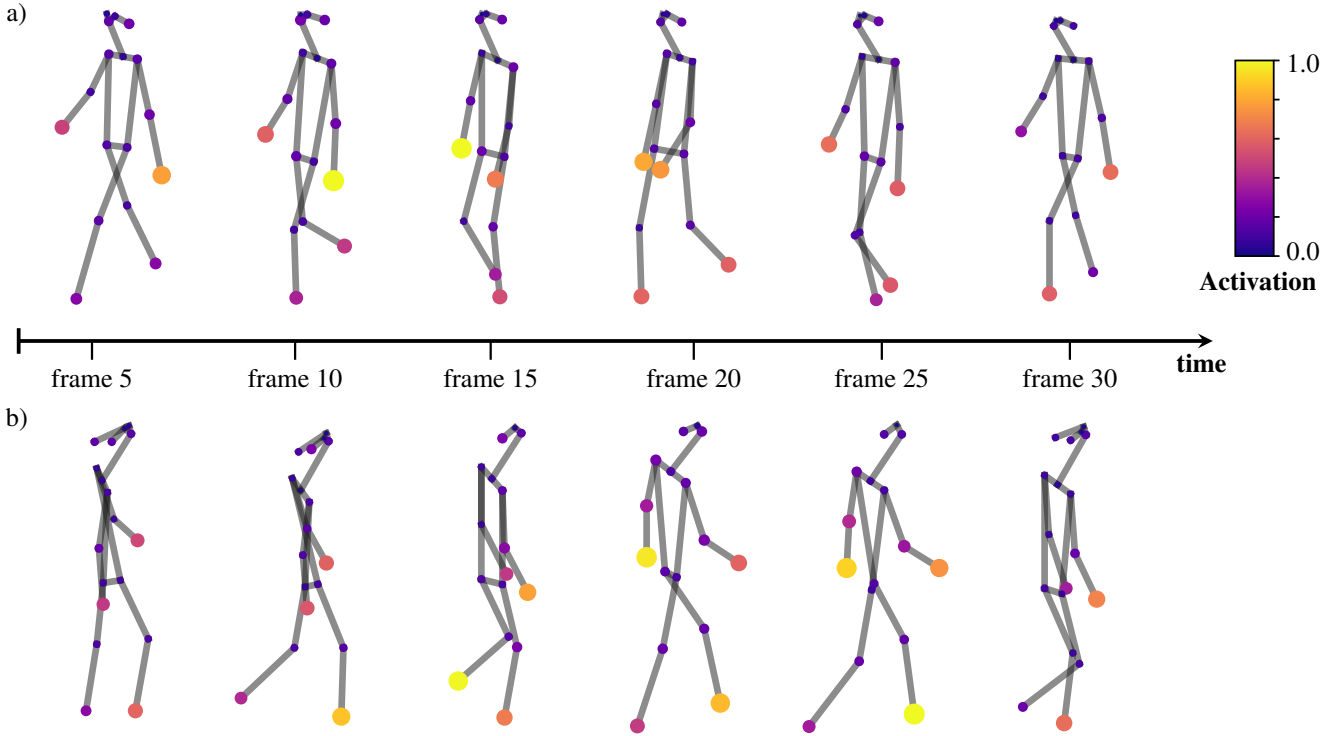
\begin{figure*}[t]
    \centering
\begin{tikzpicture}
    \draw (0, 0) node[inner sep=0] {\includegraphics[width=0.95\textwidth]{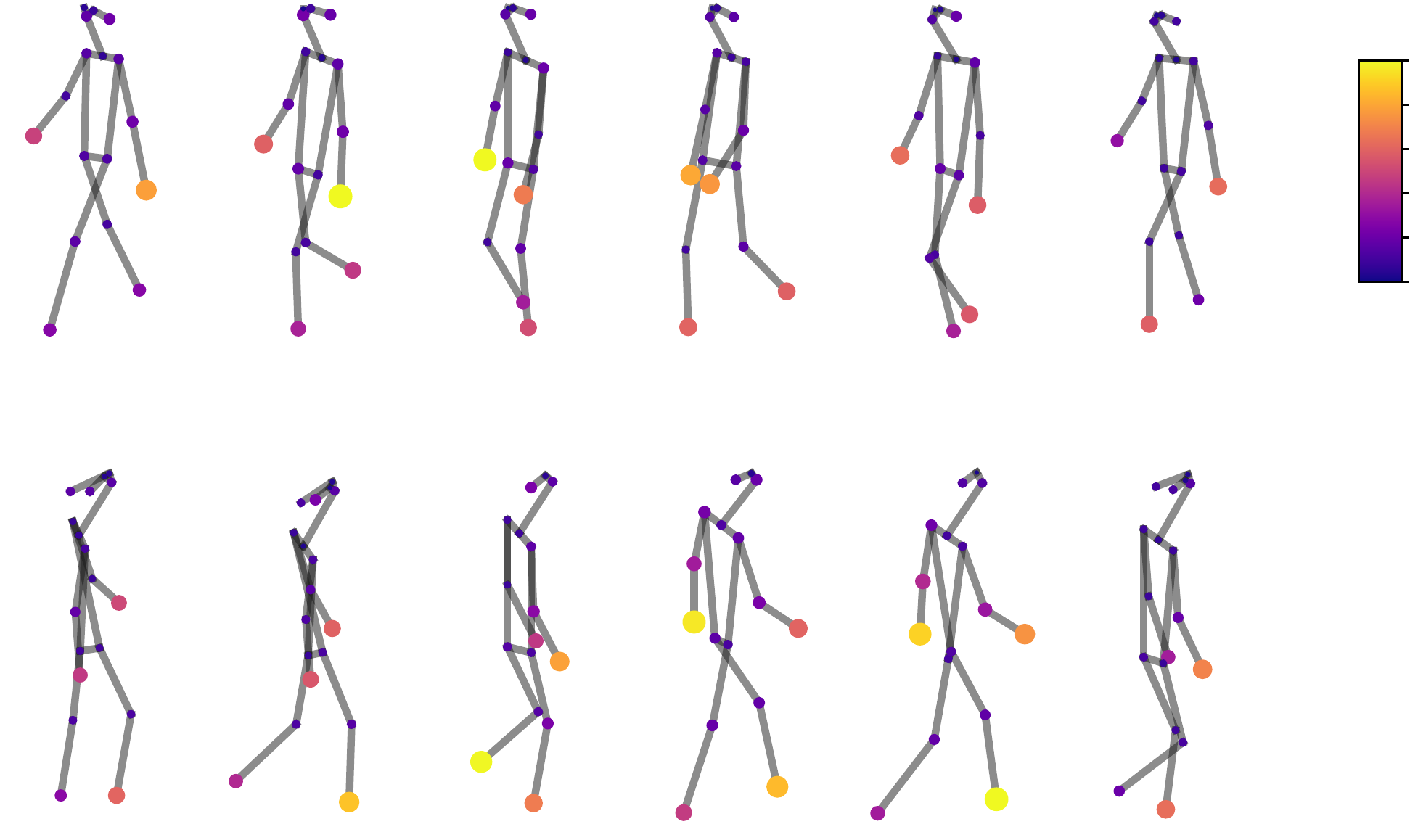}};
    \draw (-8.5, 4.85) node {a)};
    \draw (-8.5, -0.55) node {b)};

    \draw (8, 1.3) node {\textbf{Activation}};
    \draw (8.6, 4.2) node {1.0};
    \draw (8.6, 1.7) node {0.0};
    
    \draw [|-stealth, line width=0.5mm](-8.5,0.5) -- (7.4,0.5);
    \draw (7.25, 0.2) node {\textbf{time}};
    
    \draw [line width=0.3mm] (-7.5, 0.5) -- (-7.5,0.25);
    \draw [line width=0.3mm] (-4.8, 0.5) -- (-4.8,0.25);
    \draw [line width=0.3mm] (-2.2, 0.5) -- (-2.2,0.25);
    \draw [line width=0.3mm] (0.5, 0.5) -- (0.5,0.25);
    \draw [line width=0.3mm] (3, 0.5) -- (3,0.25);
    \draw [line width=0.3mm] (5.5, 0.5) -- (5.5,0.25);
    
    \draw (-7.5,0) node {frame 5};
    \draw (-4.8,0) node {frame 10};
    \draw (-2.2,0) node {frame 15};
    \draw (0.5,0) node {frame 20};
    \draw (3,0) node {frame 25};
    \draw (5.5,0) node {frame 30};
\end{tikzpicture}
\caption{Activated joints of our proposed method on an example gait cycle from the OUMVLP-Pose test set. Higher activated joints are shown in a brighter color and in bigger scale.
The view angle for a) is 60\degree and b) is 255\degree.}
\label{fig:activation}
\end{figure*}

\subsection{Gait Recognition Analysis}
A better understanding of what the network learns from the skeleton data is indicated by the discriminate features of gait. Hence we want to study which joints have the highest activation at different times of the gait cycle. Using the activation map technique  \cite{zhou2016learning} we calculate the activation of each joint per time frame as shown in \cref{fig:activation}.

The Figure shows that the network looks mainly on the outer limbs. The keypoints on the hands and feet have the highest activation over the gait cycle, and the shoulders and face keypoints (except ears) have the lowest activation. The limb with forward motion has a higher activation than fixed parts in the temporal context. For example in sequence a), the right leg swing from frame 5 to frame 20 shows increased foot activation the more it moves. These observations conclude that the arm-swing and the leg-swing are the most discriminate features for our network. The network also takes hints from the hip and head movement. For the head, the ear keypoints are most relevant, presumably because they are most outward and the most reliable keypoints detected from the back view. This observation is highly aligned with an early definition of gait "the motion of the living body [is] represented by a few bright spots describing the motions of the main joints" \cite{johansson1973visual} and the intuitive understanding of gait, which is best captured by the outer limbs.

 \section{Conclusion}
In this paper, we present a improved approach for skeleton-based gait recognition and introduce a multi-branch architecture for gait recognition. The architecture uses pre-computed temporal information divided into three branches: joints, motion, and bones. We archive impressive performance gains on OUMVLP-Pose, the largest skeleton-based gait dataset, with an improve of 3$\times$ over to the baseline. Compared to previous graph-based approaches, our training is much faster and much more stable. We are the first approach to publish results on both popular gait datasets and set the baseline for future gait recognition research.

In our ablation, we analyze the learned gait representation of the network and show our strong temporal modeling. 
Furthermore, our visualization of joint activations indicate the strong focus on moving body parts and the outermost joints. Indicating to focus on these joints for further research.
Together these two observations confirm that our model captures \textit{real} gait features, instead of performing a visual re-identification. Combined with the advantages of robust pose estimation, this allows for broader spectrum of applications of gait recognition. Our ablation indicates that gait recognition will improve with more accurate pose estimators. This closes the gap for bringing gait recognition from the lab to the real-world.

%%%%%%%%% REFERENCES
{\small
\bibliographystyle{ieee_fullname}
\bibliography{egbib}
}

\end{document}